\documentclass[conference]{IEEEtran}
\IEEEoverridecommandlockouts
\usepackage{cite}
\usepackage{amsmath,amssymb,amsfonts}
\usepackage{algorithmic}
\usepackage{graphicx}
\usepackage{textcomp}
\usepackage{xcolor}

\usepackage{booktabs}  
\usepackage{tablefootnote}
\usepackage{threeparttable}
\usepackage{multirow}
\usepackage{url}

\def\BibTeX{{\rm B\kern-.05em{\sc i\kern-.025em b}\kern-.08em
    T\kern-.1667em\lower.7ex\hbox{E}\kern-.125emX}}
\begin{document}

\title{The Few-shot Dilemma: Over-prompting Large Language Models\\
}

\author{\IEEEauthorblockN{Yongjian Tang\textsuperscript{1,2}, Doruk Tuncel\textsuperscript{1}, Christian Koerner\textsuperscript{1}, Thomas Runkler\textsuperscript{1,2}}
\IEEEauthorblockA{\textsuperscript{1}\textit{Siemens AG} \\
\textsuperscript{2}\textit{Technical University of Munich}
}

}

\maketitle

\begin{abstract}
Over-prompting, a phenomenon where excessive examples in prompts lead to diminished performance in Large Language Models (LLMs), challenges the conventional wisdom about in-context few-shot learning. To investigate this few-shot dilemma, we outline a prompting framework that leverages three standard few-shot selection methods -- random sampling, semantic embedding, and TF-IDF vectors -- and evaluate these methods across multiple LLMs, including GPT-4o, GPT-3.5-turbo, DeepSeek-V3, Gemma-3, LLaMA-3.1, LLaMA-3.2, and Mistral. Our experimental results reveal that incorporating excessive domain-specific examples into prompts can paradoxically degrade performance in certain LLMs, which contradicts the prior empirical conclusion that more relevant few-shot examples universally benefit LLMs. Given the trend of LLM-assisted software engineering and requirement analysis, we experiment with two real-world software requirement classification datasets. By gradually increasing the number of TF-IDF-selected and stratified few-shot examples, we identify their optimal quantity for each LLM. This combined approach achieves superior performance with fewer examples, avoiding the over-prompting problem, thus surpassing the state-of-the-art by 1\% in classifying functional and non-functional requirements.
\end{abstract}


\section{Introduction}
\label{section_introduction}
Instruction-tuned LLMs have demonstrated exceptional language understanding and knowledge inference capabilities \cite{zhao2024surveylargelanguagemodels}. 
Compared with fine-tuning, LLM-based few-shot prompting avoids the need for extensive training and requires minimal annotated data. 
Such streamlined problem-solving features have promoted more accessible solutions in various domains \cite{goswami-etal-2023-switchprompt}.

However, while LLMs continue to advance the performance in general benchmarks, their capabilities to acquire domain-specific knowledge from in-context examples remain ambiguous. Previous studies have encountered various constraints: token limits of earlier LLMs prevent the inclusion of adequate few-shot examples \cite{10298329,bolucu-etal-2023-impact}; outdated few-shot selection methods underutilize the few-shot learning capabilities of modern LLMs \cite{milios-etal-2023-context}; and some studies focus solely on generic benchmarks rather than domain-specific tasks \cite{wang2023gptnernamedentityrecognition}. 
Considering these limitations, we investigate few-shot prompting 
with three few-shot selection methods and multiple LLMs. 
In the experiments, we discover a few-shot dilemma caused by over-prompting, where an excess of domain-specific examples can actually degrade the performance of certain LLMs.
\begin{figure}[!t]
    \vspace{-0cm}
    \includegraphics[width=1\linewidth]{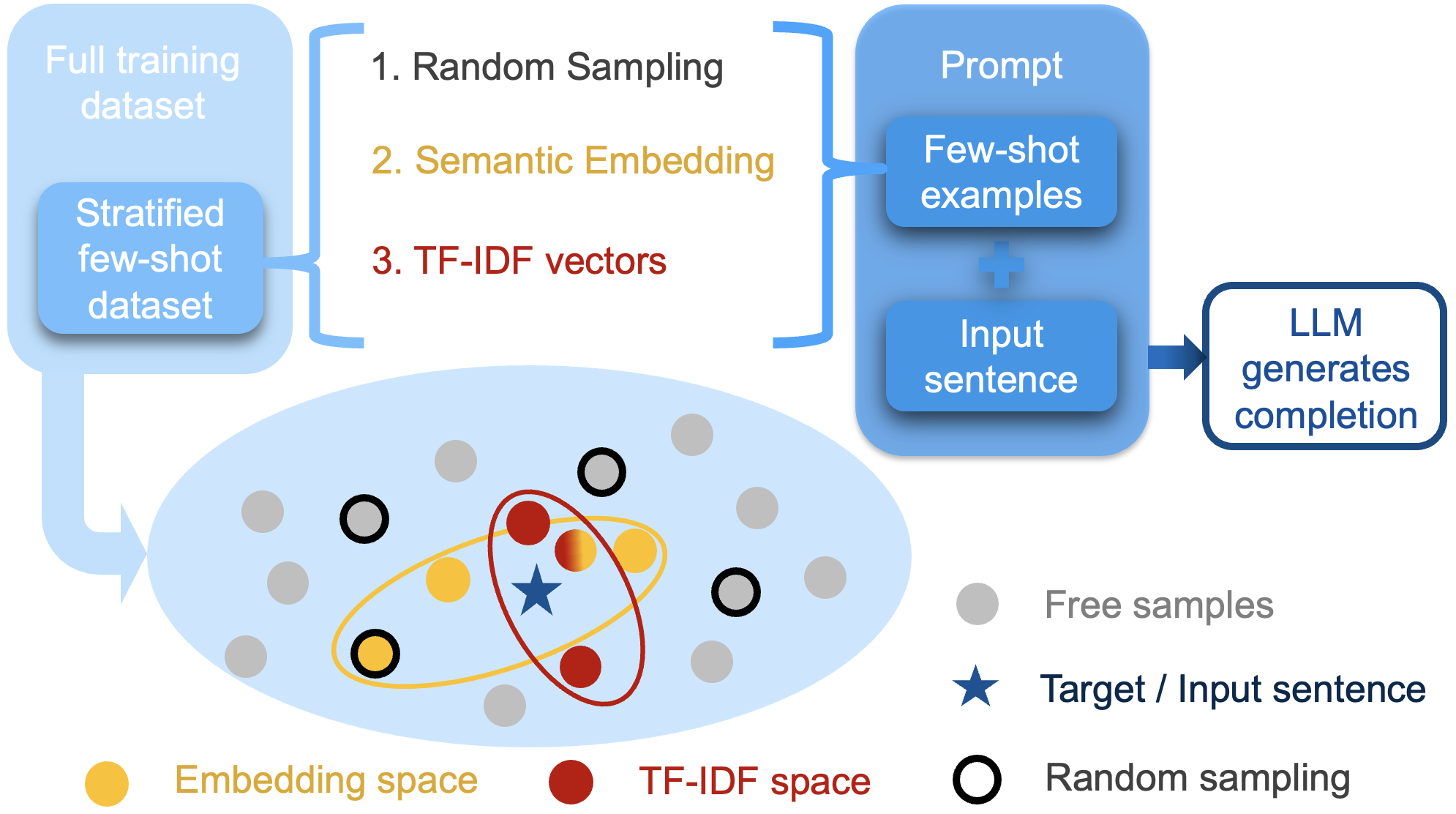}
    \centering
    \vspace{-0.3cm}
    \caption{A prompting framework with three standard few-shot selection methods. Given a target sentence, we identify the closest examples in embedding space or TF-IDF vector space. 
}
    \vspace{-0.6cm}
    \label{method_overview}
\end{figure}

Modern software systems have become increasingly complex and interconnected. 
Recent advances in LLMs have shown promising potential to transform the software engineering process, particularly in managing requirements. 
Building on this trend, we study the few-shot dilemma with two requirement classification datasets: PROMISE~\cite{li2014non} and PURE~\cite{ferrari2017pure}, aiming to automate the categorization of software requirements based on their characteristics and functionalities. 
Prior works have extensively experimented with these two datasets. 
Initially, decision trees~\cite{abad2017works} and support vector machines~\cite{kurtanovic2017automatically} were leveraged, which require complex feature engineering beforehand. 
A deep learning approach~\cite{rahman2023pre} combined pre-trained word embeddings with CNN, LSTM, and ANN architectures to enhance feature extraction and classification accuracy. Furthermore, pre-trained language models have advanced the state-of-the-art performance progressively~\cite{norbert,9920081,prcbert}.  
A recent study guided ChatGPT to classify requirements in zero-shot and few-shot settings~\cite{el2023ai}, integrating two random examples per class into the prompt. However, they do not exploit the potential of few-shot prompting thoroughly, and a comparison between prompted LLMs and fine-tuned models such as BERT variants is still missing on the two requirement classification datasets.
Given the current research gap, we include multiple instruction-tuned LLMs of varying sizes and prompt them with methodically selected few-shot examples. Our experiments unveil the few-shot dilemma -- LLM performance peaks with a proper number of examples in specific domains and then gradually declines with excessive examples, with the decline margin varying across different LLMs. To the best of our knowledge, we are the first to investigate this phenomenon caused by over-prompting through concrete experiments in software engineering domain. The main contributions of this paper can be summarized as follows:

\begin{itemize}
    \item [(i)]  
    We identify the few-shot dilemma caused by over-prompting for certain LLMs, challenging the conventional wisdom about LLM-based few-shot learning with our experimental results on two software engineering datasets.        
    \item [(ii)] 
    We propose a few-shot selection framework and investigate the few-shot dilemma by incorporating an increasing number of examples selected through random sampling, semantic embedding, and TF-IDF vectors. Notably, TF-IDF outperforms the other two methods in filtering relevant few-shot examples.
    \item [(iii)] 
    We further analyze the dynamics and rules of few-shot prompting across seven LLMs -- GPT-4o, GPT-3.5-turbo, DeepSeek-V3, Gemma-3, LLaMA-3.1, LLaMA-3.2, and Mistral, revealing model-specific optimal few-shot settings and performance characteristics.

    \item [(iv)] 
    By gradually incorporating more TF-IDF-selected few-shot examples, we identify their optimal quantity for each LLM. This approach achieves better performance with fewer examples, effectively mitigating the over-prompting effect. As a result, we surpass the state-of-the-art by 1\% in classifying functional and non-functional requirements of the PROMISE dataset, highlighting the importance of balancing example quality and quantity in LLM applications.

\end{itemize}

\section{Related Work}
\subsection{Transformers and Language Models}
With the advent of transformer architecture~\cite{vaswani2017attention}, language modeling has received significant attention~\cite{zhao2024surveylargelanguagemodels}. 
Among all models, BERT~\cite{Devlin2019BERTPO} initially stands out 
and its ”pre-training and fine-tuning” mode has inspired numerous follow-up works, such as RoBERTa~\cite{liu2019roberta}, PRCBERT~\cite{prcbert}, and NoRBERT~\cite{norbert}. 
Meanwhile, the research community continues to improve the performance of language models by scaling up their sizes. Compared with smaller counterparts, large-scale models demonstrate unseen emergent abilities~\cite{wei2022emergent} and have provided us with more problem-solving possibilities, such as prompt engineering and in-context learning. 

\subsection{In-context Learning}
In-context learning enables LLMs to generate answers directly based on the provided information without any parameter updates.
A study finds that LLMs perform optimally when accessing relevant information at the beginning or end of the input context, but their performance degrades significantly for information in the middle~\cite{liu2023lost}. When presented with conflicting in-context examples, LLMs can temporarily disregard pre-trained knowledge, and the ability to override semantic priors enhances as the model size increases. While smaller models cannot flip predictions and follow contradictory labels, larger models can achieve this~\cite{wei2023larger}. In line with this discovery, another study reveals that prompting with examples that contain irrelevant problem descriptions but correct answers actually improves the robustness of model outputs~\cite{shi2023largelanguagemodelseasily}.

\subsection{Few-shot Prompting}
\subsubsection{Order of Few-shot Examples}
\label{subsubsection_order_examples}
The order in which few-shot examples are permutated in prompts can impact performance, ranging from the state-of-the-art to a random guess~\cite{lu2021fantastically}. Incorporating additional few-shot samples into prompts can lead to considerable performance improvements, but it does not significantly reduce variance. Generally, larger models tend to achieve better performance with lower variance. 
Liu et al.~\cite{liu2024let} suggest to rank examples from simple to complex, gradually increasing complexity to improve inference performance, while another group \cite{li2024longcontextllmsstrugglelong} reveals that a scattered distribution of labels throughout the prompt outperforms class-based grouping of examples. However, few-shot selection and filtering criteria have not been explored in these studies.

\subsubsection{Selection of Few-shot Examples}
A commonly used strategy is to select high-quality few-shot examples that are semantically close in embedding space. One initial work~\cite{liu2021makes} leverages RoBERTa~\cite{liu2019roberta} as a sentence encoder to convert original sentences into embedding vectors. Based on these vectors, they apply the $k$NN algorithm to identify the \(k\) nearest examples for each input sentence. 
A follow-up study~\cite{wang2023gptnernamedentityrecognition} enhances embedding-based few-shot selection by replacing RoBERTa with SimCSE~\cite{gao2022simcse}, a contrastive learning framework developed for sentence embedding, to encode few-shot examples and input sentences.

FsPONER~\cite{tang2024fsponer}, a TF-IDF-based few-shot selection framework, outperforms the embedding-based methods in domain-specific entity recognition tasks. Rather than focusing on semantic meaning, it considers the frequency of keywords within a corpus to fairly select relevant examples for each input sentence.

\subsubsection{Rules of Few-shot Prompting}

Inspired by the properties of GPT-3~\cite{brown2020languagemodelsfewshotlearners}, it is assumed that including more demonstration examples in prompts can help LLMs understand tasks better and produce more accurate outputs in the desired formats. Furthermore, extensive few-shot examples may show variability in the domain, thereby reducing the ambiguity. However, due to the limited token length of early models, lengthy prompts were not allowed. Consequently, the research community focused on improving the quality and relevance of selected examples~\cite{liu2021makes,wang2023gptnernamedentityrecognition}.

Two recent studies have made similar observations regarding the newly released LLMs with extended context windows. In datasets featuring expansive label spaces across diverse domains, long-context models exhibit continuous performance gains with hundreds of demonstrations~\cite{bertsch2024incontextlearninglongcontextmodels}. When presented with a series of TF-IDF vector-selected examples in industrial domains, the performance of LLMs improves consistently~\cite{tang2024fsponer}, but the trend slows down once sufficient examples have been incorporated.

\section{The Few-shot Prompting Framework}
\begin{figure}[!h]
    \vspace{-0cm}
    \centering
    \includegraphics[width=1\linewidth]{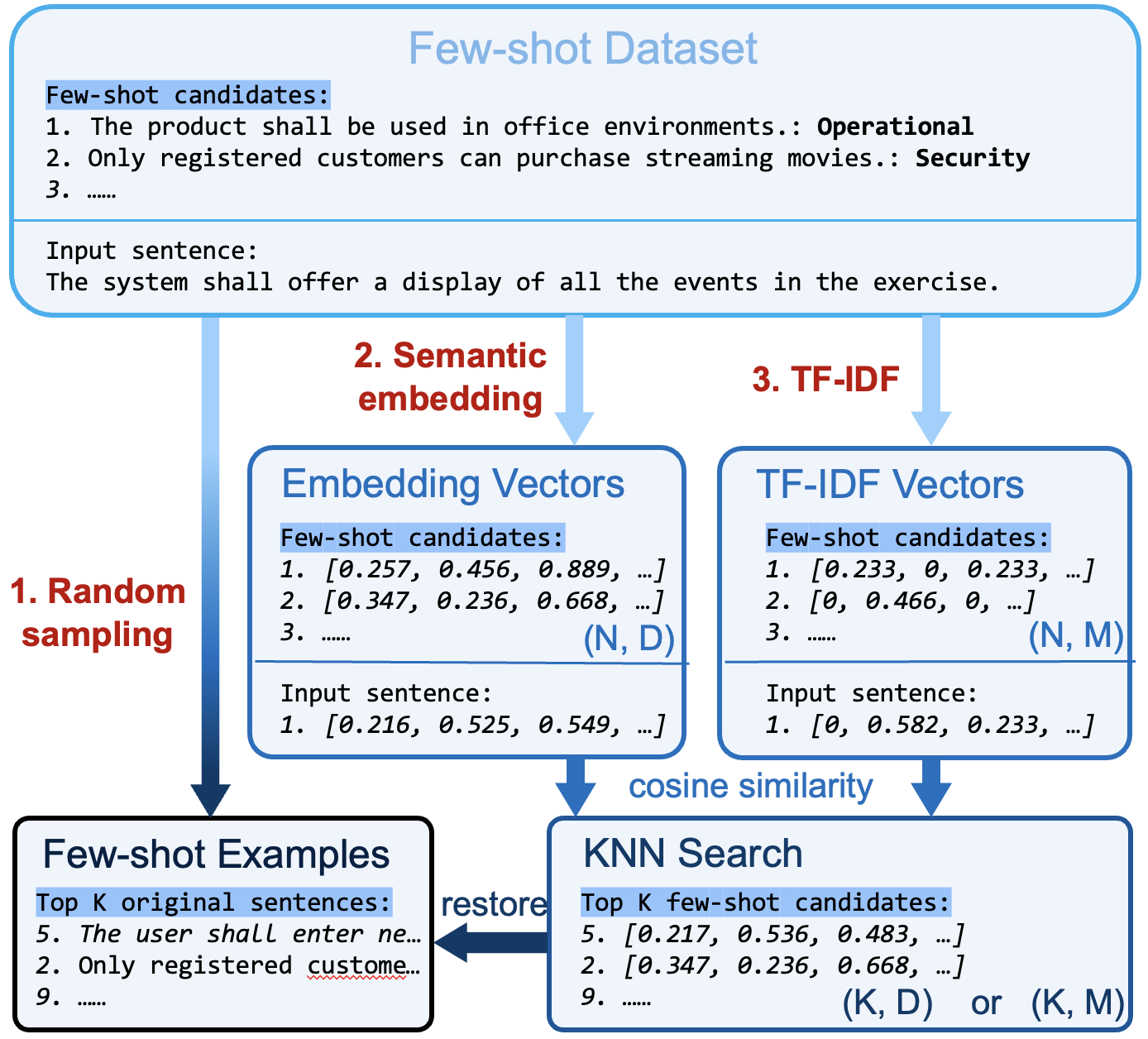}
    \vspace{-0.2cm}
    \caption{Illustration of the three few-shot selection methods used in the prompting framework. The matrix of embedding vectors is in size N times D, where N represents the total number of few-shot candidates and D denotes the dimension of embedding space. Similarly, we create the TF-IDF matrix, where N is the number of few-shot candidates and M is the quantity of individual words in this corpus.}
    \label{three_vectors}
\end{figure}

To systematically evaluate and compare the effectiveness of different few-shot selection strategies, we outline a prompting framework based on three standard few-shot selection methods: random sampling, semantic embedding, and TF-IDF vectors, as illustrated in Figures~\ref{method_overview} and ~\ref{three_vectors}.

\subsection{Few-shot Dataset Stratification}
In the PROMISE and PURE datasets, the class distribution is imbalanced. Randomly selecting few-shot examples from the original dataset may result in an overemphasis on those more common classes, potentially ignoring the less frequent ones. To address this issue, we iterate through all classes and sequentially select one example for each class. Depending on the specified quantity of few-shot examples, we determine the number of iteration rounds. This stratification approach ensures an adequate representation of small classes, enabling LLMs to learn more detailed information from the few-shot examples. Notably, a small few-shot dataset aligns with real-world constraints, where data annotation is expensive and only limited data are available.

\subsection{Few-shot Selection Methods}

\subsubsection{Random Sampling}
Random sampling is an intuitive approach where samples are randomly drawn from the few-shot dataset and used as demonstrations. Although this method is straightforward and easy to understand, it overlooks the semantic relation between sentences. Therefore, we integrate the other two methods to filter more relevant examples, as shown in Figure~\ref{three_vectors}.

\subsubsection{Semantic Embedding}
For semantic embedding, various sentence encoders can be incorporated to transform original sentences into embedding vectors, such as Sentence BERT~\cite{Reimers2019SentenceBERTSE} and SimCSE~\cite{gao2022simcse}. Based on these vectors, we compute the cosine similarity between the input sentence and all few-shot candidates, and then select the \(k\) nearest examples for prompting.
The results in this paper are derived from the few-shot examples selected with SimCSE.

\subsubsection{TF-IDF Vectors}
The TF-IDF-based method resembles the structure of semantic embedding~\cite{tang2024fsponer}, but instead of using a sentence encoder, a TF-IDF converter is integrated to transform input sentences and few-shot candidates into vectors. Based on cosine similarity, the \(k\) nearest candidates for each input sentence are selected as few-shot examples.

\subsection{Software Requirement Classification} 
Software projects continue to grow in complexity and scale. A promising approach to ensure traceability and systematic management in software engineering is requirement classification, which provides a structured way to organize requirements based on their characteristics and attributes.
In this process, the classifier takes textual requirements as inputs and generates pre-defined categories as outputs, including functional and non-functional classes in the PURE dataset \cite{ferrari2017pure}, or further divided non-functional subclasses in the PROMISE dataset \cite{li2014non}, such as performance, operational, and security. 
To tackle this problem, traditional approaches rely on deep learning models or machine learning algorithms, which require complex feature engineering and massive annotated data. More recent efforts leverage language models and their prior knowledge in the software engineering domain, achieving state-of-the-art performance with minimal labeled data.

\subsection{Prompt Structure}

\begin{figure}[!h]
    \vspace{-0cm}
    \includegraphics[width=0.95\linewidth]{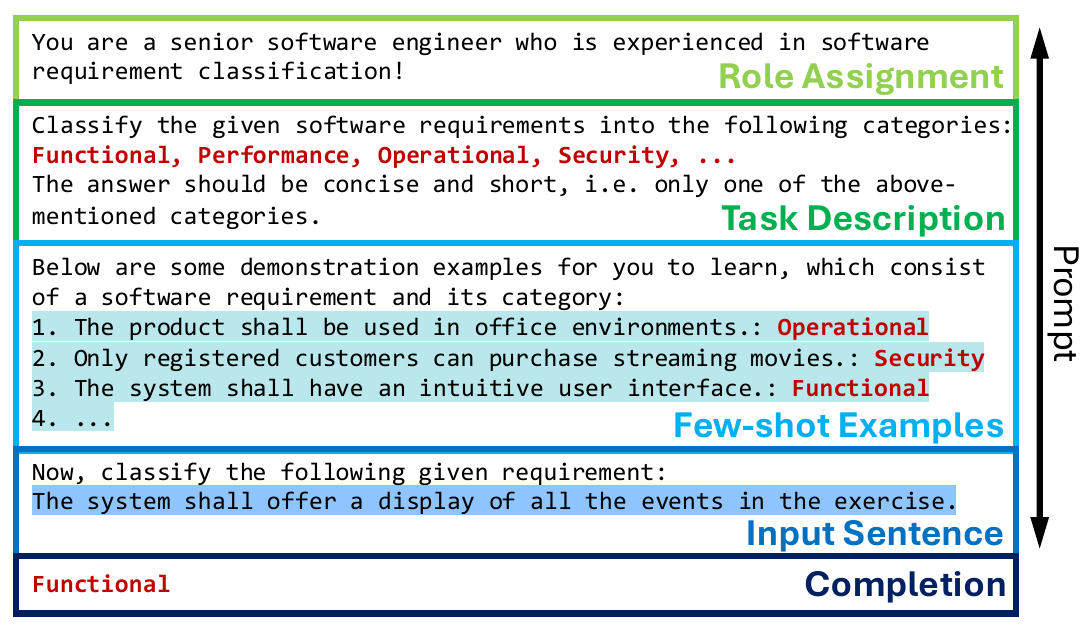}
    \centering
    \vspace{-2mm}
    \caption{The prompt used for classifying requirements consists of a system message with role assignment, a task description with target classes, selected few-shot examples, and the input sentence to be classified.}
    \label{prompt_structure}
\end{figure}

Given the software requirement classification use case, we define the prompt structure in Figure~\ref{prompt_structure}. We first assign LLMs a role, steering them to exploit the domain-specific knowledge acquired in pre-training. Subsequently, we describe the task, present the target classes, and define the required completion format. In the third block, we list all selected few-shot demonstration examples for LLMs. Finally, we add the input sentence and perform the inference based on the provided prompt.

\section{Experiments}

\subsection{Experimental Setting}
\subsubsection{Selected Models} 
We consider seven instruction-tuned LLMs in this work. Table~\ref{tab:LLM_comparison} provides an overview of these models based on their sizes, context window, input modalities, training data volume, and openness to the public. Due to our limited compute resources, the three large-scale models -- GPT-4o~\cite{openai2024gpt4technicalreport,openai2024gpt4ocard}, GPT-3.5-turbo~\cite{brown2020languagemodelsfewshotlearners}, and DeepSeek-V3~\cite{deepseekai2025deepseekv3technicalreport} are only accessible via APIs. Their performance stands for the forefront of existing LLMs, which allows us to exploit the full potential of prompting methods. Furthermore, we include Gemma-3-4B~\cite{google2025gemma}, LLaMA-3.1-8B-instruct~\cite{dubey2024llama3herdmodels} and LLaMA-3.2-3B-instruct~\cite{llama32}, three widely used and lightweight open-source models from Google and Meta AI, as well as Mistral-7B-instruct~\cite{jiang2023mistral7b}, known for surpassing Llama-2-13B on all benchmarks. 



\begin{table}[htpb]
  \scalebox{0.9}{
  \hspace{-0.6cm}
     .\begin{threeparttable}
      \setlength{\tabcolsep}{3pt}
      \begin{tabular}{l l l l l l}
        \textbf{Models} & Size &  Context & Modalities  & Training data &Open-\\
           &  & window & (input) & volumn & ness\\
        \midrule
        \textbf{GPT-4o} & - & 128k & text/image & - & close \\ 
         &  & (tokens) & audio/video &  &  \\
        \textbf{GPT-3.5-turbo} & 175B & 16k & text-only & 300B tokens &  close 
        \medskip \\
        \textbf{DeepSeek-V3} & 671B & 128k & text-only & 14.8T tokens &open
        \medskip \\
        \textbf{Gemma-3-4B} & 4B & 128k & text/image & 4T tokens + distill & open 
        \medskip \\
        \textbf{Mistral-7B} & 7B & 32k & text-only & - & open  \vspace{-0.1cm}\\
        \textbf{-instruct}  &  &  &  &  &  
        \smallskip\\
        \textbf{LLaMA-3.1}  & 8B & 128k & text-only & 15T tokens & open\vspace{-0.1cm}\\
        \textbf{-8B-instruct} &  &  & &  & 
        \smallskip\\
        \textbf{LLaMA-3.2} & 3B & 128k & text-only & distill LLaMA-3.1& open \vspace{-0.1cm}\\
        \textbf{-3B-instruct}  &  &  &  &  & 
      \end{tabular} 
      \end{threeparttable}
      }  
  \centering
  \caption[Example table]{Basic information of selected LLMs.}\label{tab:LLM_comparison}
  \vspace{-0.2cm}
\end{table}

Alongside these models, we include three BERT~\cite{Devlin2019BERTPO} variants as baselines -- RoBERTa~\cite{liu2019roberta}, NoRBERT~\cite{norbert}, and PRCBERT~\cite{prcbert}, which hold the state-of-the-art performance on the PROMISE dataset. Since the three models do not demonstrate instruction-following abilities, they must be fine-tuned for requirement classification in a supervised fashion.

\subsubsection{Selected Datasets} 
Following the trend of using LLMs to streamline the software engineering process, we study the few-shot prompting features with two datasets in this domain.
The PROMISE dataset~\cite{li2014non}, originally annotated according to ISO-25010~\cite{iso25010_old}, has served as a benchmark for requirement classification for years, with extensive prior evaluations~\cite{
rahman2023pre,han2023improving,khan2023non,abad2017works,norbert,prcbert}. The PURE dataset comprises 79 raw requirement documents and 34,268 sentences~\cite{ferrari2017pure}, with 6118 sentences annotated as functional or non-functional requirements in a recent study~\cite{sonali2024fr}.

\begin{table}[htpb]
  \scalebox{0.9}{
      \begin{threeparttable}
      \setlength{\tabcolsep}{5pt}
      \begin{tabular}{l l l l l l }
        \textbf{Datasets} & \textbf{Number}&\textbf{Fitting}&\textbf{Number}\\
         & \textbf{of words}& \textbf{tasks}&\textbf{of classes}\\ 
        \midrule
        Original PROMISE&12,000+&Binary / multi-class&12\\ 
        \vspace{-0.4cm} 
        \smallskip\\
        Relabeled PROMISE&12,000+&Binary / multi-class & 9 \\
        \vspace{-0.4cm} 
        \smallskip\\
        PURE 
        &90,000+&Binary&2\\
      \end{tabular} 
      \end{threeparttable}
  }  
  \centering
  \vspace{2mm}
  \caption[Example table]{Three selected software engineering datasets.}\label{tab:dataset_comparison}
  \vspace{-0.2cm}
\end{table}

\subsubsection{Relabeled PROMISE Dataset} 
\begin{figure}[!h]
    \vspace{-0cm}
    \hspace{-0.25cm}
    \includegraphics[width=1\linewidth]{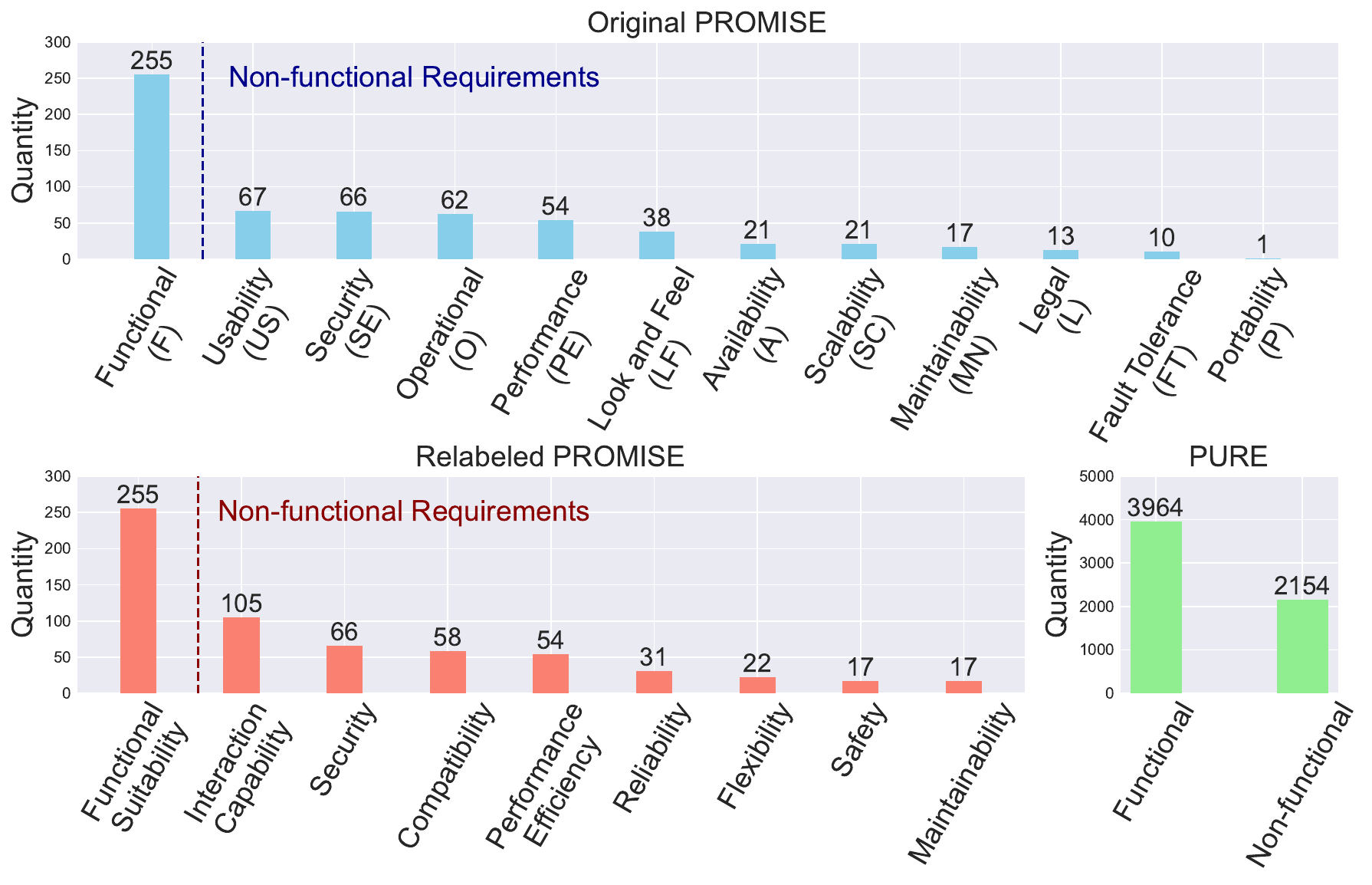}
    \centering
    \vspace{-0.2cm}
    \caption{The distribution of classes in datasets.}
    \label{distribution_datasets}
\end{figure}

To assess LLMs' capability in state-of-the-art professional requirement analysis and their understanding of software standards, we relabel the PROMISE dataset conforming to the latest version of ISO-25010~\cite{iso25010}. Moreover, we exclude lower-level sub-characteristics, such as "Look and Feel" and "Fault Tolerance", focusing on the nine high-level software characteristics to reduce ambiguity. For unchanged characteristics in the new ISO/IEC standard -- Security, Performance, and Maintainability, we maintain the original annotations. We present detailed information of the three datasets in Tab.~\ref{tab:dataset_comparison} and illustrate the distribution of classes in Fig.~\ref{distribution_datasets}.

\subsection{Binary Classification on Functional and Non-functional Requirements}
We first evaluate the selected LLMs on classifying functional and non-functional requirements on both PROMISE and PURE datasets with an 80-20 split. 

\subsubsection{Comparison of Few-shot Selection Methods}

\begin{figure}[!h] 
    \vspace{-0cm}
    \hspace{-0.1cm}
    \vspace{-0.1cm}
    \includegraphics[width=1\linewidth]{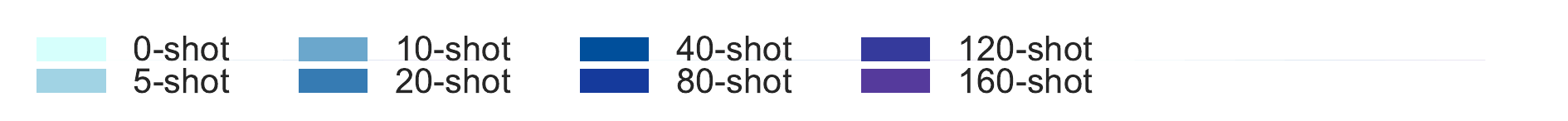}
    \centering
    \includegraphics[width=1\linewidth]{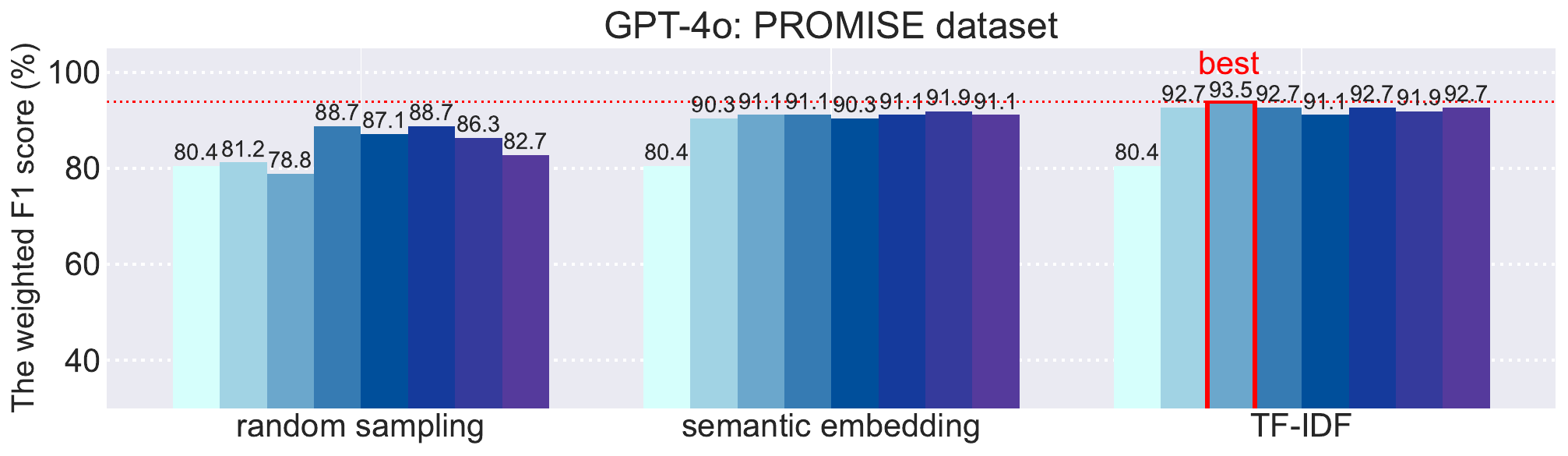}
    \centering
    \includegraphics[width=1\linewidth]{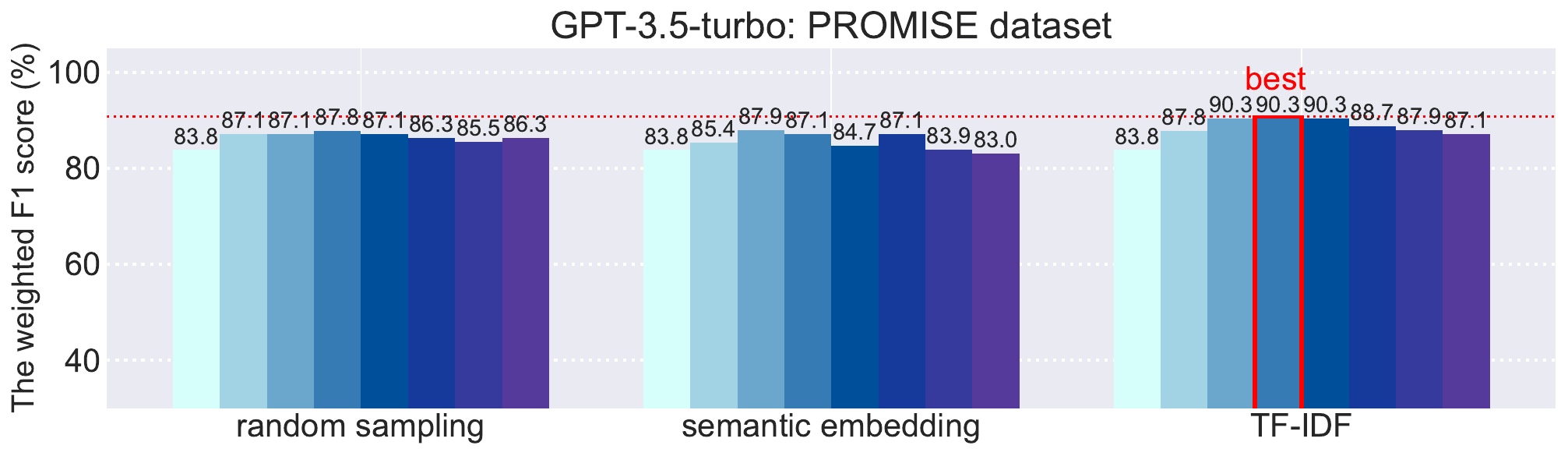}
    \centering
    \includegraphics[width=1\linewidth]{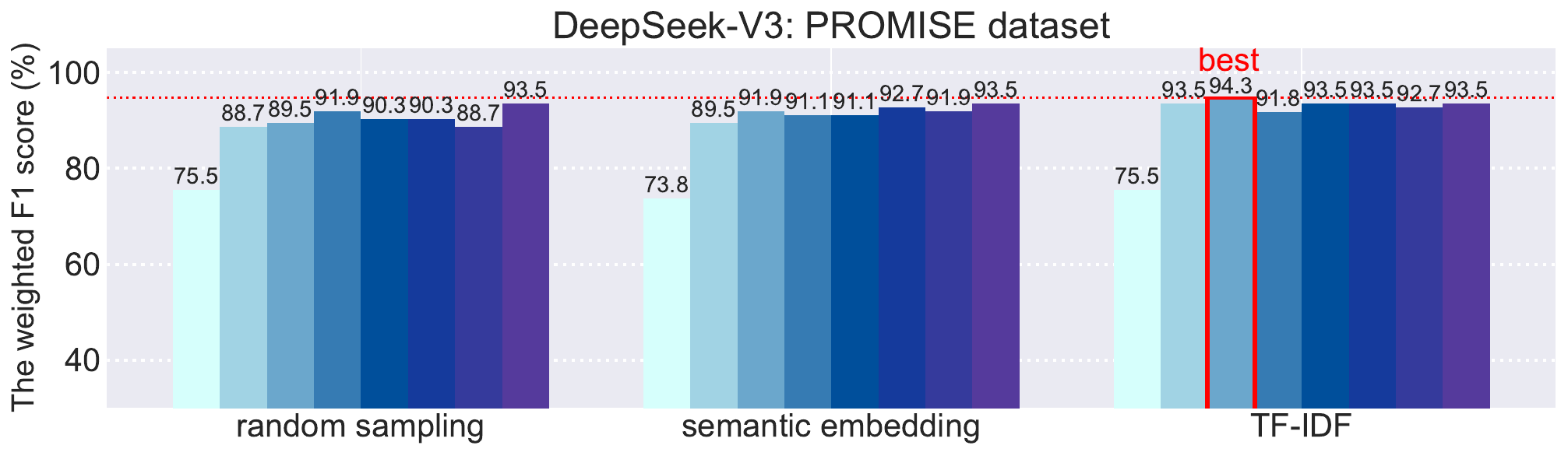}
    \centering
    \includegraphics[width=1\linewidth]{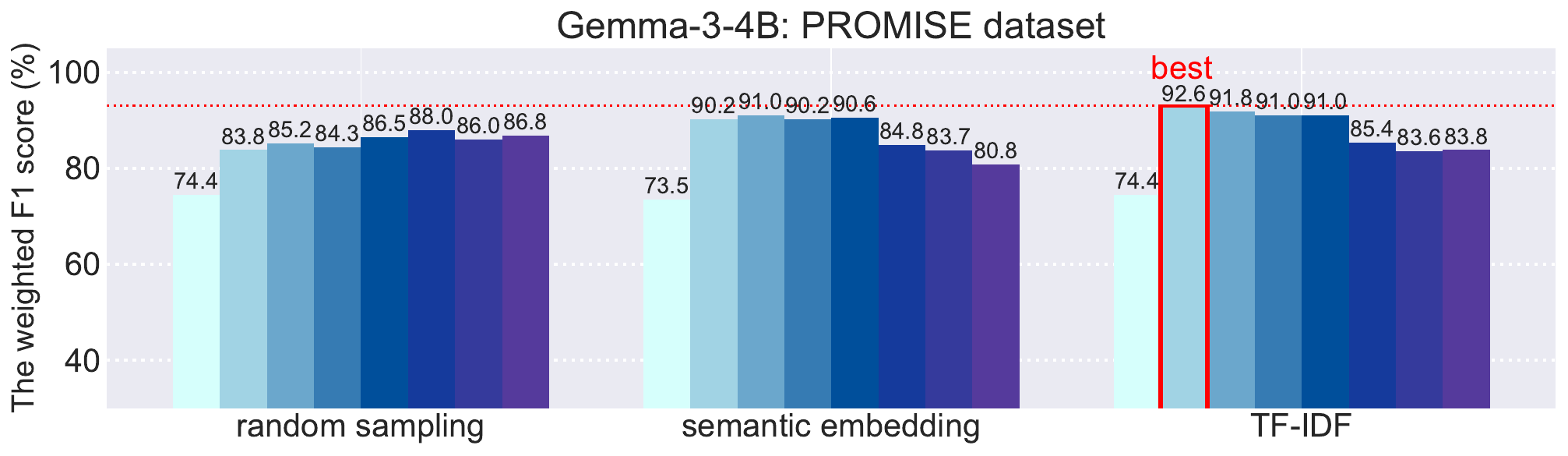}
    \centering
    \includegraphics[width=1\linewidth]{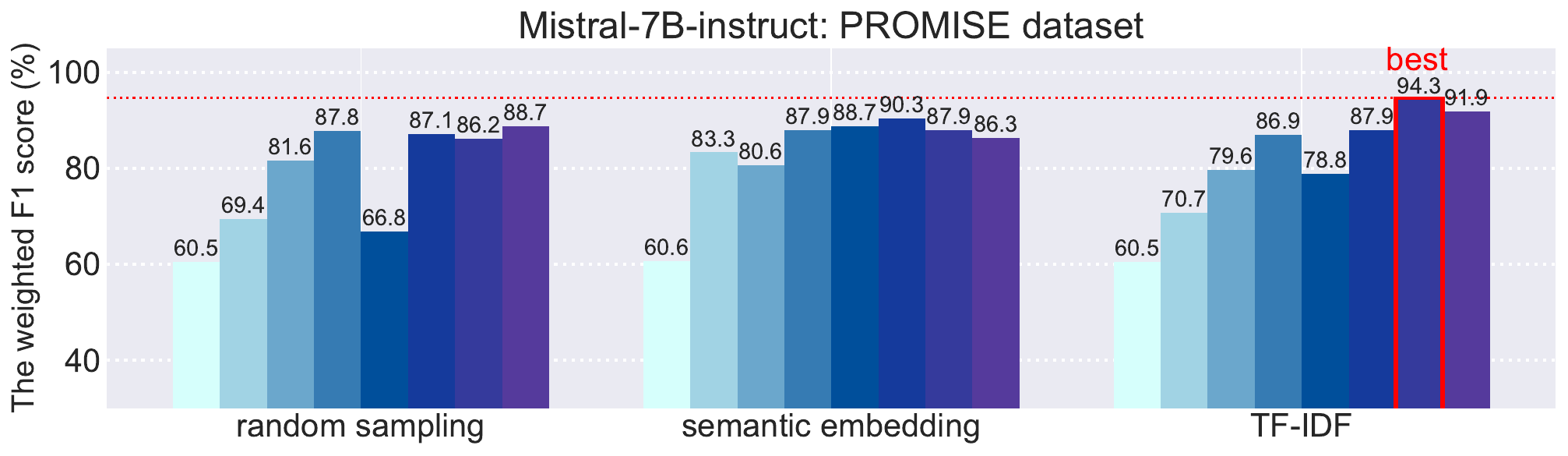}
    \centering
    \includegraphics[width=1\linewidth]{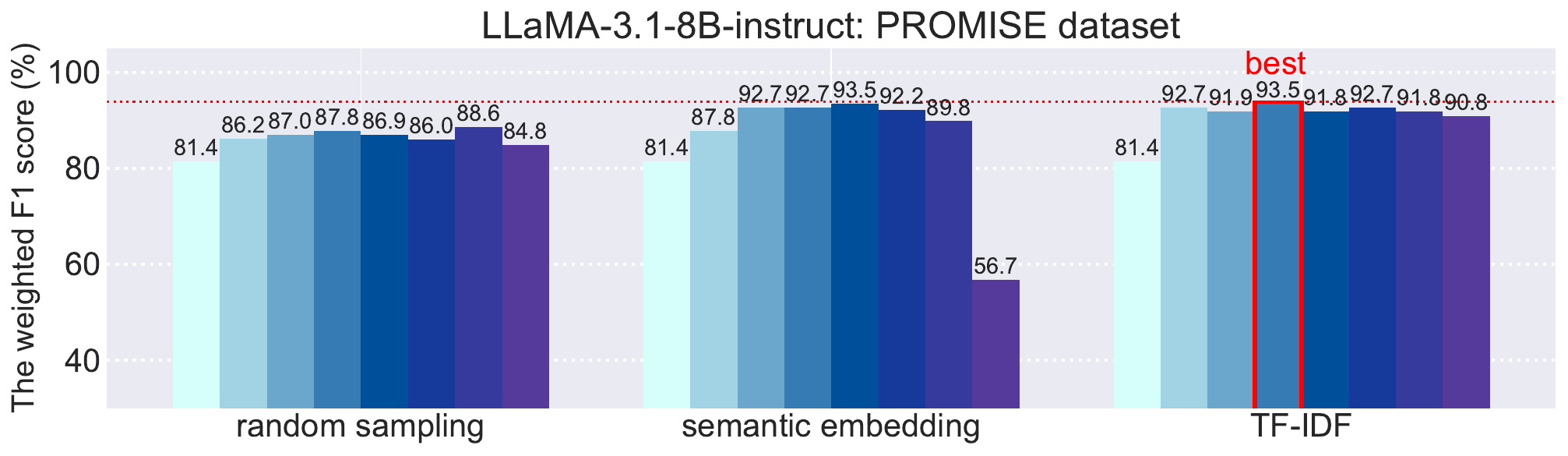}
    \centering
    \includegraphics[width=1\linewidth]{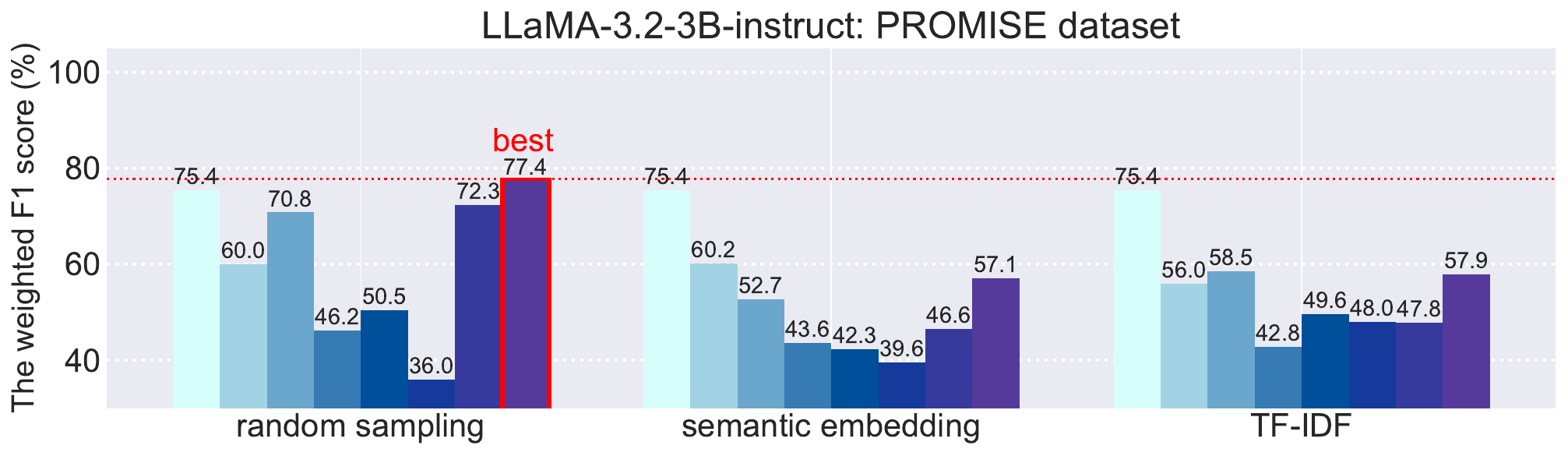}
    \centering
    \vspace{-3mm}
    \caption{The comparison among three few-shot selection methods on the original PROMISE dataset for all models, with the X-axis representing the selection methods and the Y-axis indicating the weighted F1 score. The color gradient, from light to dark, reflects an increasing number of few-shot examples ranging from 0 to 160 (see the legends above the figure). In total, our binary evaluation on this dataset involves 168 experiments, i.e. 7(models) $\times$ 3 (methods) $\times$ 8 (few-shot quantities).}
    \vspace{-0.2cm}
    \label{bar_graph_binary_1}
\end{figure}

As shown in Figure~\ref{bar_graph_binary_1}, 
almost all models achieve their optimal performance with TF-IDF-based few-shot selection. The only exception -- LLaMA-3.2-3B-instruct performs best with random sampling. The model struggles to comprehend long contexts when presented with more than 20 examples, resulting in an output that is either exclusively functional classes or random guesses. 
We also studied the impact of few-shot order -- shuffling examples during random sampling, but the average performance remains inferior to stratified semantic embedding and TF-IDF vectors.
To further validate the effectiveness of the latter two methods, we evaluated all models on the larger PURE dataset and attained similar results. 

\subsubsection{Observed Over-prompting Behavior}
\begin{figure}[!h]
    \hspace{-0.2cm}
    \includegraphics[width=1\linewidth]{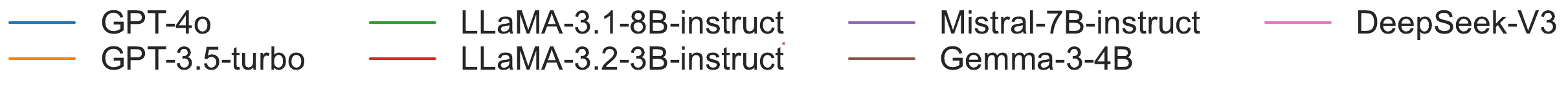}
    \includegraphics[width=1\linewidth]{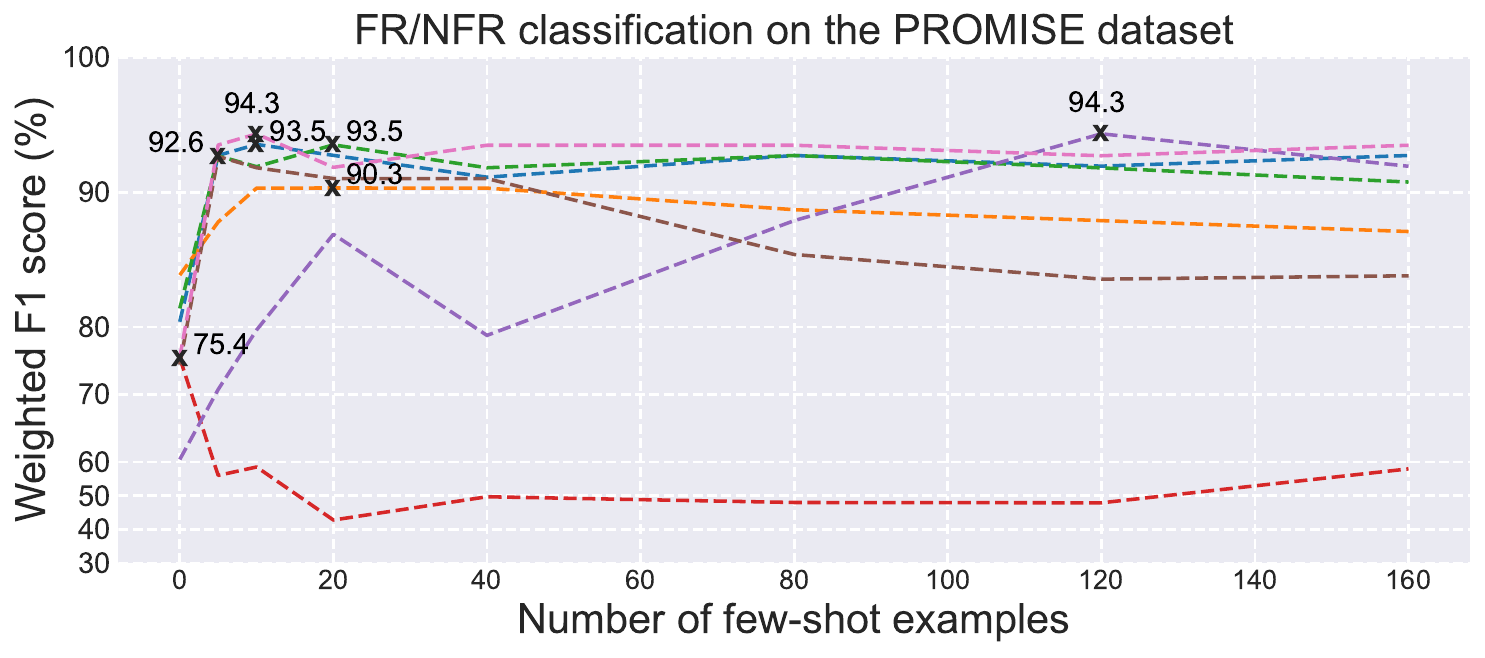}
    \centering
    \includegraphics[width=1\linewidth]{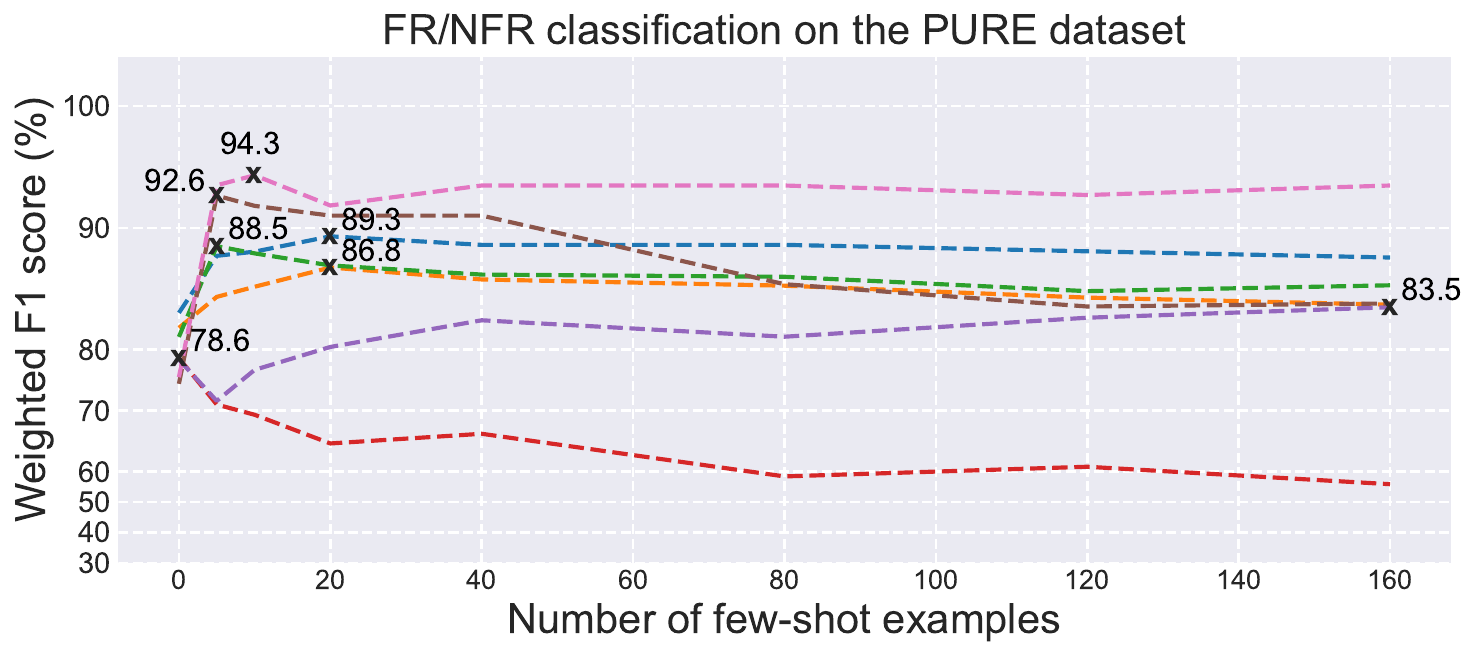}
    \centering
    \vspace{-0.6cm}
    \caption{Performance variation of all models on the PROMISE and PURE datasets with an increasing number of few-shot examples selected by TF-IDF vectors.}
    \vspace{-0cm}
    \label{line_graph_binary_1}
\end{figure}

Figure~\ref{line_graph_binary_1} illustrates the line graphs of all models while using TF-IDF based selection. 
As the number of few-shot examples increases, we observe the effects of over-prompting in both datasets, where the performance escalates to a high point and begins to drop slowly once sufficient examples have been integrated into the prompts. This observation deviates from the existing conclusions that performance improves or plateaus as more few-shot examples are added~\cite{bertsch2024incontextlearninglongcontextmodels,tang2024fsponer}.

Over-prompting is more evident in the larger PURE dataset. After achieving optimal performance with approximately 5 to 20 examples, the performance of GPT-4o, GPT-3.5-turbo, LLaMA-3.1-8B, and Gemma-3-4B gradually declines, while DeepSeek-V3 reaches the plateau. The smallest model, LLaMA-3.2-3B, exhibits weakening performance from the beginning, likely due to its limited capacity to comprehend long contexts. In contrast, Mistral-7B learns consistently with more few-shot examples, potentially due to the sliding window attention mechanism~\cite{child2019generating,beltagy2020longformer} used in training, which exploits transformer's stacked layers to extend attention beyond the window size to previous contexts.

\subsubsection{State-of-the-art Comparison}
Given the peak performance points in Figure~\ref{line_graph_binary_1} and the corresponding number of few-shot examples, we obtain the optimal few-shot setting for each LLM and compare their performance to the state-of-the-art in Table~\ref{tab:binary_comparison}. 
To reduce variance and obtain more robust evaluation results, we replace the 80-20 split of data with a 10-fold cross-validation.
\begin{table}[htpb]
  \centering
  \hspace{-0.1cm}
  \scalebox{0.85}{
    \begin{threeparttable}
        \setlength{\tabcolsep}{3.5pt}
        \begin{tabular}{ll|ccc|ccc|c}  
            & &  \multicolumn{3}{|c|}{ \textbf{FR (255)} }  &  \multicolumn{3}{|c|}{  \textbf{NFR (370)} } & \textbf{Overall}\\  
            &\textbf{Models/Approaches}  & P & R               & F1 & P & R                    & F1  & F1 \\  
            \hline
            \multirow{6}{*}{ \textbf{Instruct-} }&GPT-4o                   & .85 & \textbf{.99} & .92 & \textbf{.99} & .88 & .93& .93  \\   
            \multirow{6}{*}{ \textbf{LLMs} }&GPT-3.5-turbo            & .84 & .95 & .89 & .96 & .87 & .91& .91   \\  
            &DeepSeek-V3              & .93 & .94 & .94 & .96 & .95 & .95 & .95  \\
            &Gemma-3-4B               & \textbf{.94} & .88 & .91 & .92 & .96 & .94 & .93 \\
            &Mistral-7B-instruct      & .84 & .98 & .90 & .98 & .87& .92&  .91  \\     
            &LLaMA-3.1-8B-instruct    & .93 & .94 & \textbf{.94} & .95 & \textbf{.96} & \textbf{.96} &  \textbf{.95} \\    
            &LLaMA-3.2-3B-instruct    & .48 & .98 & .64 & .93& .24 & .38&  .48  \\     

            \hline  
            \multirow{3}{*}{ \textbf{BERT} }&PRC-RoBERTa-large & \textbf{.92} & \textbf{.94} & \textbf{.93} & \textbf{.95} & .96 & \textbf{.95} & \textbf{.94}   \\
            \multirow{3}{*}{ \textbf{variants \tnote{$\dagger$}} } &PRC-BERT-base     & .91 & .92 & .91 & .91 & \textbf{.96} & .93 & .92 \\
            &NoR-BERT-large     & .92 & .88 & .90 & .92 & .95 & .93 & .92  \\
            &NoR-BERT-base     & .89 & .88 & .90 & .92 & .95 & .93 & .92  \\
            \hline  
            \multirow{3}{*}{ \textbf{ML} } & SVM (autom. features)            & .88 & .87 & .87 & .87 & .88 & .87 & .87   \\
            \multirow{3}{*}{ \textbf{algorithms} } & SVM (hand-picked features)\tnote{*}& \textbf{.92} & .93 & .93 & .93 & .92 & .92 & .92  \\
            & C4.5 DT (unprocessed data)      & .84 & .93 & .88 & .95 & .88 & .91 & .90   \\
            & C4.5 DT (processed data)\tnote{*}  & .90 & \textbf{.97} & \textbf{.93} & \textbf{.98} & \textbf{.93} & \textbf{.95} & \textbf{.94}          
            \vspace{0.1cm}                                  \\
        \end{tabular}
        \begin{tablenotes}\scriptsize
        \item [$\dagger$] Source of BERT variants \cite{liu2019roberta,norbert,prcbert}
        \item [*] Support Vector Machines (SVM) \cite{kurtanovic2017automatically} and C4.5 Decision Trees (DT) \cite{abad2017works} incorporate additional manual feature selection in data processing.
        \end{tablenotes}
    \end{threeparttable}
    }  
    \vspace{0.2cm}
    \caption[Example table]{FR/NFR classification on the PROMISE dataset (10-fold cross validation).}\label{tab:binary_comparison}
    \vspace{-0.6cm}
\end{table}

By incorporating the most relevant examples from TF-IDF-based selection, LLaMA-3.1-8B-instruct avoids over-prompting with fewer examples, outperforming all evaluated LLMs with a weighted F1 score of 95\%. 
In comparison, GPT-4o achieves the highest recall of 99\% for functional requirements, but its lower precision suggests a tendency to misclassify non-functional requirements as functional. The imbalanced class distribution in the dataset explains this discrepancy.

\begin{table*}[t!]
  \centering
  \hspace{-0.2cm}
  \scalebox{0.7}{
    \begin{threeparttable}
        \setlength{\tabcolsep}{4pt}
        \begin{tabular}{ll|c|ccc|ccc|ccc|ccc|ccc|ccc|ccc|ccc|ccc|ccc|c}
            & & \multicolumn{1}{|c|}{ \textbf{FR} } 
            & \multicolumn{30}{|c|}{ \textbf{NFR} } \\  
            & & & & \textbf{A} & & & \textbf{FT} & & & \textbf{L} & & &\textbf{LF} & & & \textbf{MN} & & & \textbf{O} & & & \textbf{PE} & & & \textbf{SC} & & & \textbf{SE} & & & \textbf{US} & &  \\  
            & \textbf{Models/Approaches} & \textbf{F1}& \textbf{P}&\textbf{R}&\textbf{F1}&\textbf{P}&\textbf{R}&\textbf{F1}&\textbf{P}&\textbf{R}&\textbf{F1}&\textbf{P}&\textbf{R}&\textbf{F1}&\textbf{P}&\textbf{R}&\textbf{F1}&\textbf{P}&\textbf{R}&\textbf{F1}&\textbf{P}&\textbf{R}&\textbf{F1}&\textbf{P}&\textbf{R}&\textbf{F1}&\textbf{P}&\textbf{R}&\textbf{F1}&\textbf{P}&\textbf{R}&\textbf{F1} & \textbf{Ave.} \\  
            \hline
            \multirow{6}{*}{ \textbf{Instruct-} } & 
            GPT-4o  & .92 & .51 & 1.0 & .68 & 1.0 &.40 & .57 & .90 & .83 & .87 & .73 & .79 & .76 & .40 & .59 & .48 & .55 & .39 & .45 & .83 &.56 & .67 & .69 & 43 & .53 & .88 &.80 &.84 & .78 &.87 &.82 & .76 \\  
            \multirow{6}{*}{ \textbf{LLMs} } & GPT-3.5-turbo   & .89 & .91 & .95 & .94 & 1.0& .60 &.75 & .83 & .95 & .89 & .86 &.79 & .83 & .71 & .59 & .65 & .83 & .73 & .78 & .70 & .74 &.72 & .76 & .62 & .68 & .90 & .71 & .80 & .92 & 69 &.79 & .82 \\  

            & DeepSeek-V3  & .94 & 1.0  & 1.0  & \textbf{1.0}  & .92 & .94  & \textbf{.93} & .91 & .89 & .90 & .91 & .91 & \textbf{.91} & .83 & 1.0 & \textbf{.91} & .86 & .86 & \textbf{.86} & 1.0 & .93 & \textbf{.96} & .89 & .89 & \textbf{.89} & .95 & .89& .92 & .93 & .81 & .87 & \textbf{.92} \\ 

            & Gemma-3-4B  & .91 & 1.0 & .50 & .67 & .90 & .90 & .90 & .87 & .88 & .88 & .0 & .0  & .0 & .40 & .80 & .53 & .86 & .86 & .86 & .82 & 1.0 & .90 & 1.0 & .50 & .67  & 1.0 & 1.0 & \textbf{1.0} & .93 & .72 & .81  & .77 \\ 
            
            & Mistral-7B-ins.    &.90 & .91 & .95 & .93 & 1.0 & .6 & .75 & .83 &.95 &.89 & .86& .79 &.82 &.71 & .59& .65 & .83 & .75 &.78 & .70&.74&.72 & .76&.62& .68 &.90 & .71& .80 &.92 & .69 & .79 & .81 \\ 
            & LLaMA-3.1-8B-ins.  &\textbf{.94} & .73 & .90& .81 & 1.0 & .70 & .82 & .94 & .91 & \textbf{.92} & .76 & .89& .82 & .86 &.35&.50 & .81&.81 & .81 & .82 &.91& .86 & .77& .81 & .79 & .83 &.88 & .85 & .92 &.82 &\textbf{.87} & .86\\ 
            & LLaMA-3.2-3B-ins. &.64 & .50 & .62& .55 & .50 &.10 & .17 & .91 &.83 &.85 & .82 & .61& .70 & .0 &.0 &.0 & .71&.51&.61 & .83&.37&.51 & 1.0&.38& .55 & .76&.77&.77 & .90&.69 &.78 & .71 \\ 
            \hline  
            \multirow{3}{*}{\textbf{BERT}  } &
            PRC-RoBERTa-L  & \textbf{.93} & - & - &\textbf{1.0} & - &  - & \textbf{1.0} &  - &  - & \textbf{.96} &  - &  - &  \textbf{.93} &  - &  - &  \textbf{.94} &  - &  - & \textbf{.94} &  - &  - &  \textbf{.96} &  - &  - &  \textbf{.91} &  - &  - &  \textbf{.98} &  - &  - &  \textbf{.97} &  \textbf{.96} \\
            \multirow{3}{*}{\textbf{variants\tnote{$\dagger$}}  } & PRC-BERT-B  & .91 &  - &  - &  .93 &  - &  - &  .74 &  - &  - &  .68 &  - &  - &  .90 &  - &  - &  .86 &  - &  - &  .93 &  - &  - & .91 &  - &  - &  .87&  - &  - &  .98 &  - &  - & .94  & .91 \\
            & NoR-BERT-L  & .90 & .80 &.76&.78 & .60 &.60 &.60 &.91 &.77 &.83 &.81 &.79 &.80 &.62 &.47 &.53 &.78 &.84 &.81 &.92 &.87 &.90 &.76 &.76 &.76 &.90 &.92 &.91 &.83 &.88 &.86 &.82 \\
            & NoR-BERT-B  & .90 & .77 &.81 &.79 &.60 &.30 &.40 &.91 &.77 &.83 &.80 &.74& .77 &.70& .41 &.52& .83& .84 &.83 &.94 &.87 &.90& .64 &.67 &.65 &.78& .92& .85 &.78 &.85& .81 &.80 \\
            \hline  
            \textbf{ML} & NB (DP)\tnote{*}   &-  & .90&.90&\textbf{.90} & .90&.97&\textbf{.93} & 1.0&.75&\textbf{.86} & .94&.94&\textbf{.94} &.82&.90&\textbf{.86} & .91&.78&\textbf{.84} & 1.0&.90&\textbf{.95} & .83&.83&\textbf{.83} & 1.0&.97&\textbf{.98} & .77&.97&\textbf{.86} & \textbf{.90}\\
            \textbf{algorithms} & LDA (DP)\tnote{*}  &-  & .60&.95&.74 & .02&.10&.03 & .20&.47&.28 & .85&.60&.70 & .52&.70&.60 & .70&.35&.47 & .95&.70&.81 & .57&.81&.70 & .87&.87&.87 & .76&.61&.68 & .62
            \vspace{1.5mm}\\

        \end{tabular}
        \begin{tablenotes}\smallskip
          \item[$\dagger$] Source of BERT variants \cite{liu2019roberta,norbert,prcbert}
          \item[*] 
          Latent Dirichlet Allocation (LDA) and Naive Bayes (NB) incorporate manual feature selection in data processing (DP) \cite{abad2017works}.
        \end{tablenotes}
    \end{threeparttable}
    }  
    \vspace{0.15cm}
    \caption[Example table]{State-of-the-art comparison on multi-class classification of the original PROMISE dataset (10-fold).}\label{tab:multi-class_comparison}  
    \vspace{-0.5cm}   
\end{table*}

\subsection{Multi-class Classification on the Original PROMISE Dataset}
Adhering to an analogous experimental setup, we conduct multi-class classification on the original PROMISE dataset with an 80-20 train-test split to determine the optimal few-shot setting for each LLM. We then further validate the performance of all approaches through a 10-fold cross-validation.

\subsubsection{Observed Over-prompting Behavior}

\begin{figure}[!h]
    \hspace{-0.2cm}
    \includegraphics[width=1\linewidth]{line_graph_legends.pdf}
    \includegraphics[width=1\linewidth]{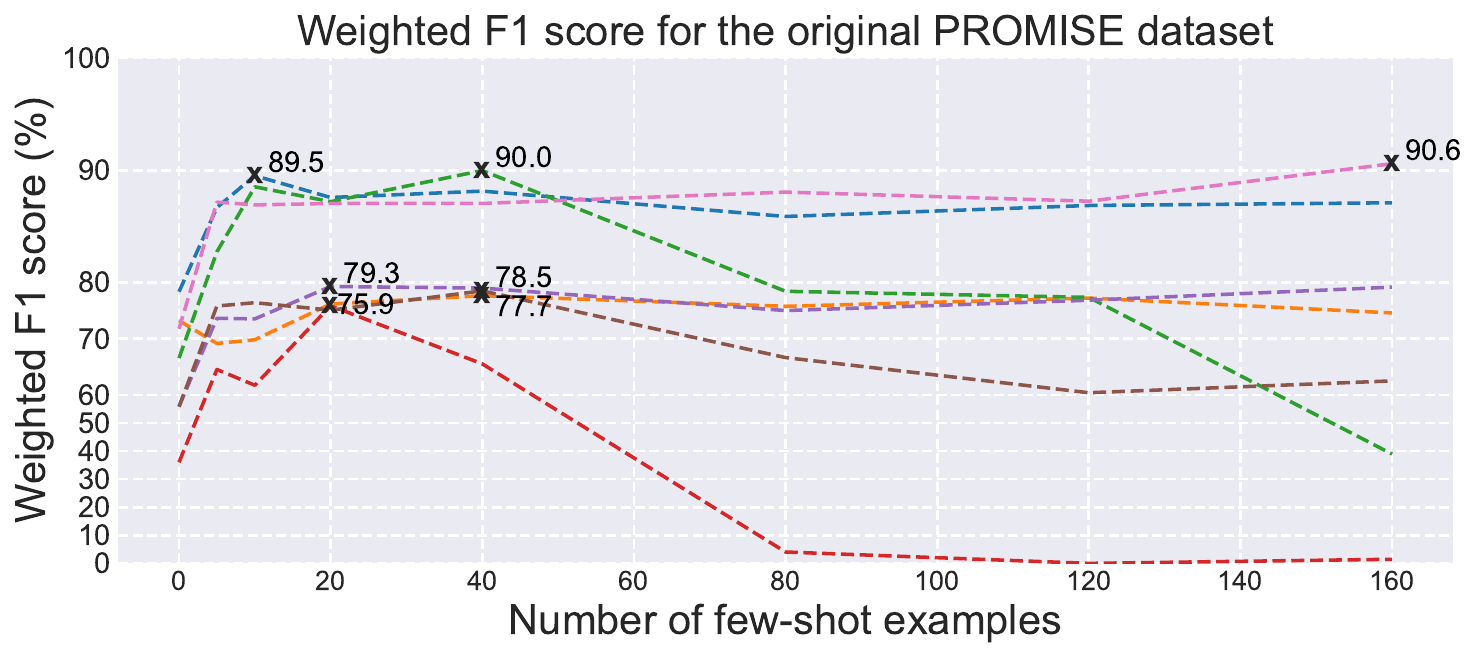}
    \centering
    \includegraphics[width=1\linewidth]{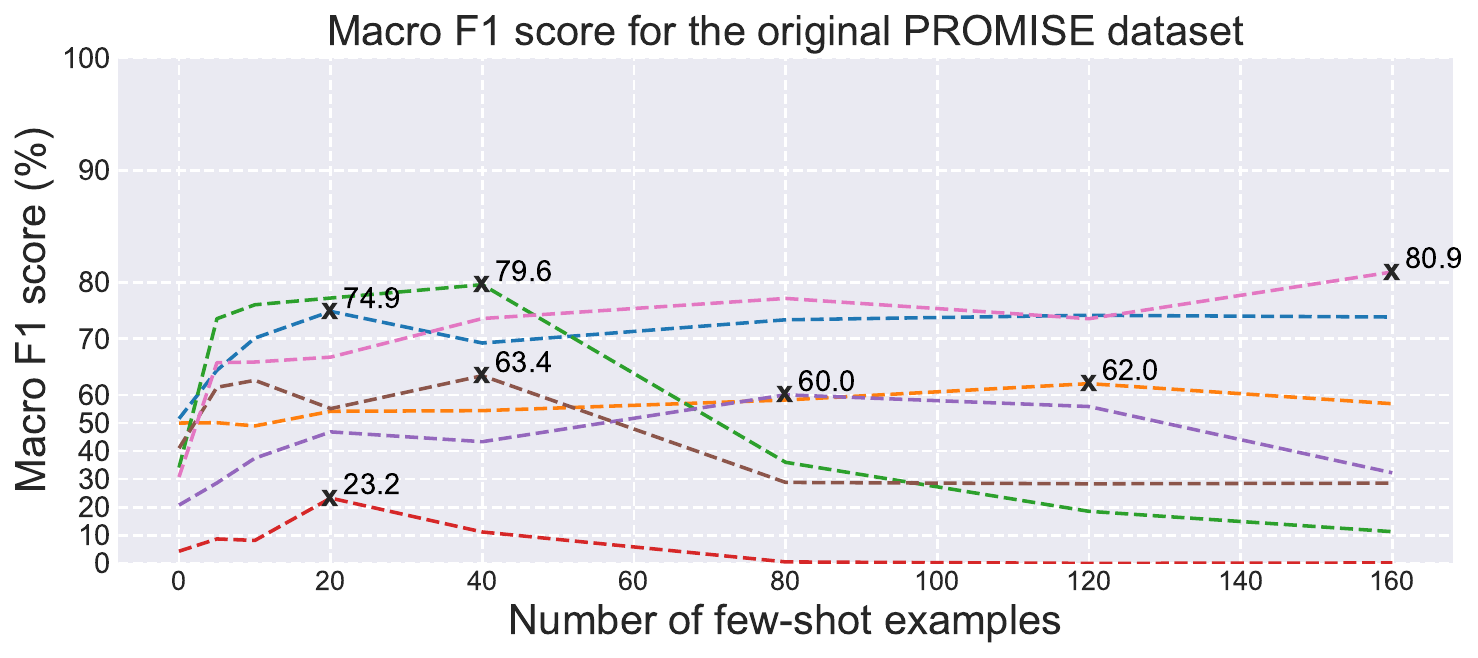}
    \centering
    \vspace{-0.7cm}
    \caption{The variation of weighted and macro F1 scores of all models on multi-class classification of the original PROMISE dataset, as the number of TF-IDF selected few-shot examples increases.}
    \label{line_graph_original_promise}
\end{figure}

Figure~\ref{line_graph_original_promise} demonstrates the weighted and macro F1 score of all models for multi-class classification. DeepSeek-V3 and LLaMA-3.1-8B-instruct stand out as the top performers, reaching weighted F1 scores of 90.6\% and 90.0\% respectively. Nevertheless, the macro F1 score is generally lower than the weighted F1 score across all models, suggesting that LLMs tend to focus on more dominant classes while undervaluing less frequent ones.

Furthermore, we observe a more apparent over-prompting phenomenon in multi-class classification. The performance of the two LLaMA models and Gemma-3-4B declines dramatically with more few-shot examples, both in weighted and macro F1 score, while another compact model, Mistral-7B-instruct, delivers consistent performance in weighted F1 score, but exhibits a hill-shaped curve in macro F1 score.
Examining the generated classes closely, LLaMA-3.2-3B tends to assign several classes to one single requirement, with performance worsening significantly beyond 80 examples.
In contrast, the two GPT models do not show a significant performance drop, and DeepSeek-V3 achieves a notable performance gain when scaling from 120 to 160 few-shot examples. The applied YaRN~\cite{pengyarn} strategy and the two-phase extension training explain its superior performance. 

\subsubsection{State-of-the-art Comparison}
Following the optimal quantity of TF-IDF-selected examples from Figure~\ref{line_graph_original_promise}, we perform 10-fold cross-validation on the PROMISE dataset and compare our approach with other state-of-the-art solutions in Table~\ref{tab:multi-class_comparison}.

DeepSeek-V3 delivers the best performance among all instruction-tuned models, achieving a weighted F1 score of 92\%, 
while GPT-4o underperforms, evidenced by an F1 score of 76\%.
One plausible hypothesis for the cause centers on the knowledge disparity arising from different information refresh windows. The PROMISE dataset was originally annotated based on the older ISO-25010~\cite{iso25010_old}, while GPT-4o was trained predominantly on more recent data. This temporal misalignment could explain the suboptimal performance of GPT-4o, motivating us to relabel the PROMISE dataset according to the latest ISO-25010 standard \cite{iso25010}.

\subsection{Multi-class Classification on the Relabeled PROMISE Dataset}

\subsubsection{Observed Over-prompting Behavior}

\begin{figure}[!h]
    \vspace{-0cm}
    \hspace{-0.2cm}
    \includegraphics[width=1\linewidth]{line_graph_legends.pdf}
    \includegraphics[width=1\linewidth]{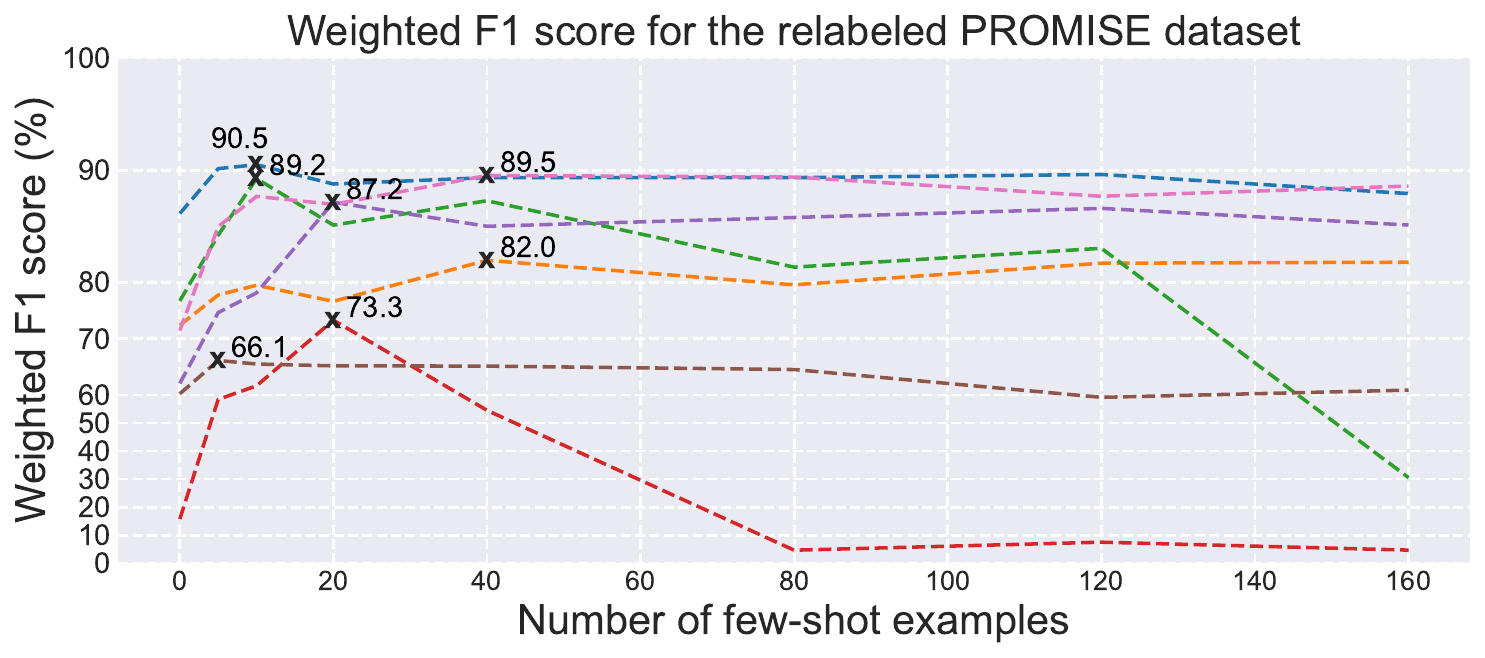}
    \centering
    \includegraphics[width=1\linewidth]{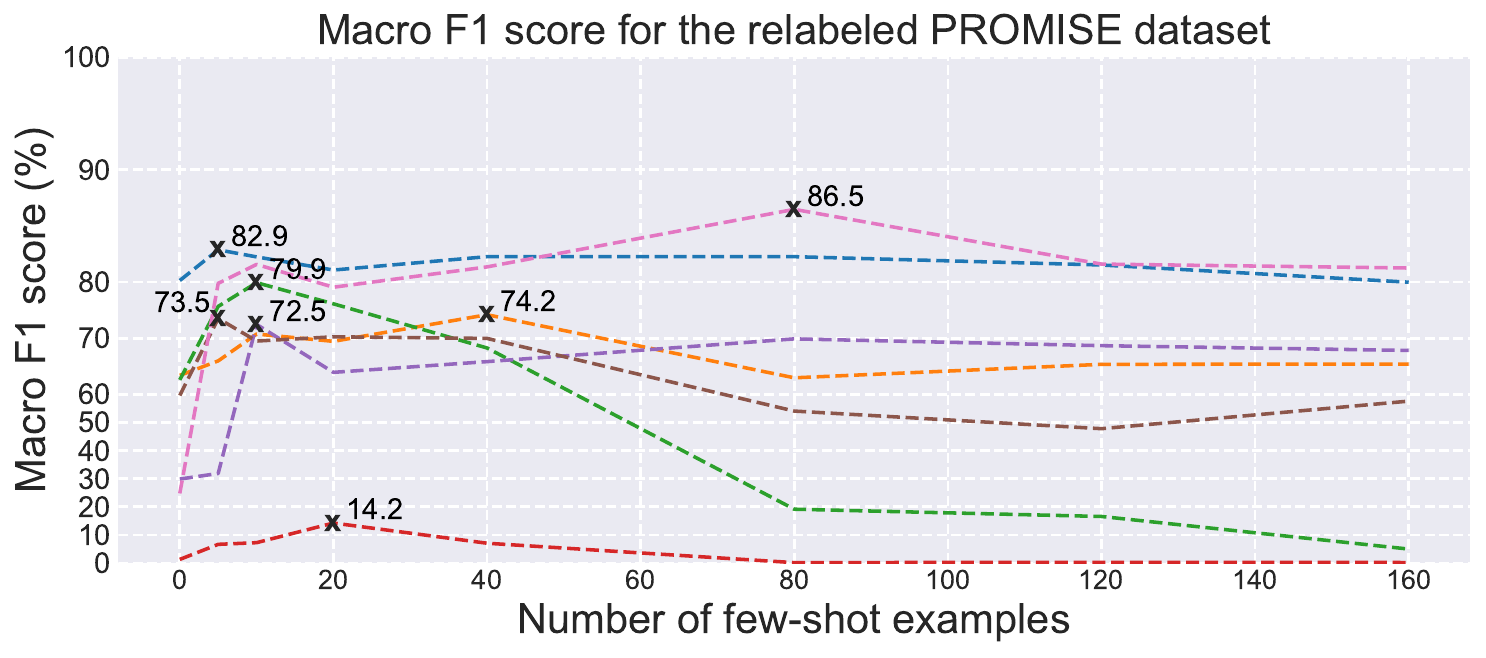}
    \centering
    \vspace{-0.6cm}
    \caption{The variation of weighted and macro F1 scores of all models on multi-class classification of the relabeled PROMISE dataset, as the number of TF-IDF selected few-shot examples increases.}
    \vspace{-0.5cm}
    \hspace{-0.3cm}
    \label{line_graph_relabeled_promise}
\end{figure}

As illustrated in Figure~\ref{line_graph_relabeled_promise}, GPT-4o outperforms the other six LLMs in weighted F1 score, reaching 90.5\% on the relabeled PROMISE dataset. The results corroborate our hypothesis on underperformance, confirming the sensitivity of LLMs to the temporal disparity of training data.

Compared with the results in Figure~\ref{line_graph_original_promise}, all LLMs attain their optimal performance with fewer examples from the relabeled dataset, with the peak points of their over-prompting curves shifting leftward on the x-axis. This may be attributed to the enhanced alignment between the relabeled high-level requirement classes and the prior knowledge of LLMs acquired in pre-training.

\section{Conclusions and Future Work}
We identified the few-shot dilemma when incorporating excessive domain-specific examples into LLMs. The optimal performance or turning point resulting from over-prompting depends on the inherent long-context comprehension capabilities of LLMs and the relevance of provided few-shot examples. We outlined three few-shot selection methods to filter relevant examples, where TF-IDF outperforms random sampling and semantic embedding in our software requirement classification use cases, achieving superior performance with fewer examples and thereby avoiding the side-effect of over-prompting. Leveraging this combined approach, the original LLaMA-3.1-8B surpasses the state-of-the-art fine-tuned PRC-BERT by 1\% in classifying functional and non-functional requirements when given 10 to 20 selected few-shot examples. 
In contrast, large-scale models, exemplified by DeepSeek-V3 and GPT-4o, exhibit more consistent performance in understanding lengthy contexts. Their superior capability to utilize expanded contextual information allows them to improve or at least maintain stability when over-prompted with selected few-shot examples. 
Furthermore, our findings suggest that 8B parameters may represent a threshold for effective few-shot comprehension, as indicated by Gemma-3-4B's declining performance and the substantial performance gap between LLaMA-3.1-8B and LLaMA-3.2-3B. 

Future work should address the computational overhead of LLMs when processing numerous few-shot examples. Moreover, as suggested by LLaMA-3.1-8B, evaluation of emerging models on various software engineering tasks may reveal efficient alternatives to current top-performing LLMs. 
Furthermore, based on our findings, whether the trending Cache-Augmented Generation (CAG)~\cite{chan2024don} works effectively across LLMs of varying sizes and architectures should be examined.

\end{document}